\documentclass{article}
\usepackage{microtype}
\usepackage{graphicx}
\usepackage{subfigure}
\usepackage{booktabs}
\usepackage{pifont}
\usepackage{multirow}
\usepackage{amsfonts}
\usepackage{amsmath}
\usepackage{longtable}
\usepackage{algorithm}
\usepackage{algorithmic}
\usepackage{bm}
\usepackage{makecell,boldline}
\newcommand{\cmark}{\ding{51}}%
\newcommand{\xmark}{\ding{55}}%
\usepackage{hyperref}

\usepackage[accepted]{icml2019}

\icmltitlerunning{Towards Accurate Model Selection in Deep Unsupervised Domain Adaptation}

\begin{document}

\twocolumn[
\icmltitle{Towards Accurate Model Selection in Deep Unsupervised Domain Adaptation}

\icmlsetsymbol{equal}{*}

\begin{icmlauthorlist}
	\icmlauthor{Kaichao You}{tsinghua,lab}
	\icmlauthor{Ximei Wang}{tsinghua,lab}
	\icmlauthor{Mingsheng Long}{tsinghua,lab}
	\icmlauthor{Michael I. Jordan}{berkeley}
\end{icmlauthorlist}

\icmlaffiliation{tsinghua}{School of Software}
\icmlaffiliation{lab}{BNRist, Research Center for Big Data, Tsinghua University, Beijing, China}
\icmlaffiliation{berkeley}{University of California, Berkeley, USA.

Kaichao You $<$youkaichao@gmail.com$>$}
\icmlcorrespondingauthor{Mingsheng Long}{mingsheng@tsinghua.edu.cn}

\icmlkeywords{Machine Learning, ICML}

\vskip 0.3in
]

\printAffiliationsAndNotice{}

\begin{abstract}
Deep unsupervised domain adaptation (Deep UDA) methods successfully leverage rich labeled data in a source domain to boost the performance on related but unlabeled data in a target domain. However, algorithm comparison is cumbersome in Deep UDA due to the absence of accurate and standardized model selection method, posing an obstacle to further advances in the field. Existing model selection methods for Deep UDA are either highly biased, restricted, unstable, or even controversial (requiring labeled target data). To this end, we propose \textit{Deep Embedded Validation} (\textbf{DEV}), which embeds adapted feature representation into the validation procedure to obtain unbiased estimation of the target risk with bounded variance. The variance is further reduced by the technique of control variate. The efficacy of the method has been justified both theoretically and empirically. 
\end{abstract}

\section{Introduction}

Deep learning enables machine recognition~\cite{cite:CVPR16DRL, long_fully_2015} at the cost of large scale labeled data. It is common to trade off the limited labeling budget against the demand for more labeled data by data-hungry deep models. Domain adaptation~\cite{cite:TKDE10TLSurvey} serves as a promising solution to such a dilemma: it transfers the knowledge from existing labeled data (\textit{source domain}) to the unlabeled data (\textit{target domain}) to reduce the labeling work.

The formulations of domain adaptation mainly fall into two categories, \textit{covariate shift} and \textit{label shift}, relating to causal and anti-causal inference~\cite{scholkopf_causal_2012}. Although some works focus on the \textit{label shift}~\cite{lipton_detecting_2018, azizzadenesheli_regularized_2019}, \textit{covariate shift} appears more natural in recognition tasks where deep models have shown their superiority \cite{cite:ICML15DAN,cite:ICML15RevGrad}.

While shallow learning methods have been extensively studied to tackle domain adaptation problems~\cite{cite:CVPR12GFK,cite:CVPR13SA}, deep models~\cite{cite:ICML15DAN, cite:ICML15RevGrad, cite:CVPR18MCD} are attracting more and more attention because of their impressive performance. Distribution matching methods~\cite{cite:ICML17JAN,cite:AAAI18WDA} align domains with well-defined statistical distribution divergence between deep features, while adversarial learning methods~\cite{cite:ICML15RevGrad, cite:CVPR17ADDA} learn domain-invariant deep representations with adversarial training~\cite{cite:NIPS14GAN}. Other approaches to deep domain adaptation  include generative models~\cite{cite:CVPR18GenerateAdapt}, similarity learning~\cite{cite:CVPR18SL}, to name a few.

\begin{table*}[htbp]
	\caption{Comparisons among different model selection methods for Deep UDA.}
	\label{tab:comparison}%
	\vskip 0.05in
	\centering
	\begin{tabular}{lcccc}
		\toprule
 \multirow{2}{140pt}{Method} & \multicolumn{2}{c}{Working Assumptions} & \multicolumn{2}{c}{Technical Advantages} \\
\cmidrule(lr){2-3} \cmidrule(lr){4-5}
 & covariate shift & w/o target labels & unbiased & controlled variance \\
\midrule
\textbf{Source Risk} & \xmark & \cmark & \xmark & \xmark \\
\textbf{Target Risk} & \cmark & \xmark & \cmark & \cmark \\
\textbf{IWCV}~\cite{cite:JMLR07IWCV} & \cmark & \cmark & \cmark & \xmark \\
\textbf{TrCV}~\cite{cite:TrCV} & \cmark & \xmark & \cmark & \xmark \\
\textbf{DEV} (Proposed)  & \cmark & \cmark & \cmark & \cmark \\
\bottomrule
	\end{tabular}%
	\vskip -0.1in	
\end{table*}%

The typical pipeline of machine learning is as follows: several hyperparameter configurations are tried to get candidate models, then model selection methods are employed to select the best configuration, and the final result in the test set is reported. Such a diagram struggles in domain adaptation: the performance in the test set (\textit{Target Risk}) is what we care about, but labels for the test set are inaccessible both in the model training and selection stage. Labels in the training set are abundant, but the performance in the training set (\textit{Source Risk}) is inconsistent to the target risk because of the domain shift, posing an obstacle to model selection in domain adaptation. Contemporarily there are several plausible hyperparameter selection methods:

\textbf{(1)} \textbf{Fixed Hyperparameters.} \citet{cite:CVPR17ADDA, cite:CVPR18MCD,cite:CVPR18SL} stated that hyperparameters are fixed across various experiments on the same dataset. However, it remains unclear how these fixed hyperparameters were selected. A reasonable hypothesis would be that hyperparameters are selected in one task with target labeled data and applied across other tasks, which requires more than one task at hand. Such a requirement may be satisfied in the research area where there are several tasks in each dataset. Nonetheless, in practical domain adaptation scenarios, we are often interested in one task, and Fixed Hyperparameters strategy will not work.

\textbf{(2)} \textbf{Source Risk.} \citet{cite:ICML15RevGrad} selected hyperparameters by taking the source risk into account. Although the source domain is related to the target domain and source risk may reflect the target risk to some extent, such a method comes without theoretical guarantees and is not convincing. Specifically, source risk is a highly biased estimator of the underlying target risk in the presence of a large domain gap. 

\textbf{(3)} \textbf{Target Risk.} \citet{cite:ICML18CYCADA} leaved a proportion of target data to be held out for model selection and the rest target data for transductive training. While it is an unbiased estimator of ground truth target risk, it is controversial to employ labeled target data in unsupervised domain adaptation. If some labeled target data are available, then instead of using them for model selection, why not exploring them for semi-supervised domain adaptation that often yields better empirical results?

\textbf{(4)} \textbf{Importance-Weighted Cross-Validation (IWCV)} \cite{cite:JMLR07IWCV}\textbf{.}
Initially designed as a validation method for the covariate shift problem, IWCV was adopted by \cite{cite:NIPS18CDAN} to tune hyperparameters. While it has a theoretical guarantee that it is unbiased, {IWCV} requires known density ratio to approximate the target distribution. If no density ratio is given, it fits a multi-dimensional normal distribution to estimate the density ratio. Furthermore, the variance of {IWCV} is unbounded, explaining its instability.

A detailed comparison of these model selection methods is presented in Table~\ref{tab:comparison}. Different domain adaptation algorithms employ different model selection methods. It is thus challenging to compare state of the art Deep UDA models if they are selected by different validation methods. For example, it is unfair to compare the performance selected by Target Risk with that selected by Source Risk. For a fair comparison, researchers may struggle to re-implement existing methods under the same validation scheme. Further, the same work may exhibit very different results due to inconsistent model selection methods in a variety of publications.

\textbf{Dilemma in VisDA Challenge.} Synthetic-to-Real Visual Domain Adaptation (VisDA)~\cite{peng_visda} is a large scale domain adaptation challenge. It aims to facilitate the development of unsupervised domain adaptation and provides the largest cross-domain object classification dataset to date. 
When it comes to model selection, however, the organizers get trapped into the dilemma: the labels of test set cannot be published and can only live in the test server; the labels of training set are given, but presented with the considerable domain gap, their effect for model selection is limited. As a compromise, the organizers released a fully labeled validation set which is different from both the source domain and target domain, only used for model selection.

To combat the above dilemma in Deep UDA, we propose \textit{Deep Embedded Validation} (\textbf{DEV}), a new model selection method tailored to Deep UDA. It embeds adapted feature representation in the validation procedure to yield unbiased estimation of the target risk with bounded variance. Control variate method is exploited to further reduce the variance of the estimation. Theoretical analysis shows the advantage of {DEV}. Furthermore, several empirical experiments show the performance of models selected by {DEV} approaches that of {Target Risk}, even though no target labeled data are required.

\section{Related Work}

In this section, we review the model selection methods in both supervised learning and domain adaptation. 

\subsection{Model Selection in Single Domain}
\label{sec:validation}
Machine learning algorithms aim to learn knowledge from data. While learning is carried out using the \textit{training set}, our interest mainly focuses on the performance of the algorithm on unseen data, which gives rise to the usage of the \textit{test set}. Test data is held out in the training stage. It is tested only once. Hyperparameters of machine learning algorithms are selected on another part of data dubbed as the \textit{validation set}.

\textbf{Hold-out}. If abundant data is given, we can just split the data into three parts. The \textit{training set} is only for learning, the \textit{validation set} is only for hyperparameter tuning, and the \textit{test set} is for the final evaluation. We call such a method hold-out because the \textit{validation set} is not involved in training. In hold-out validation, each candidate model will only be evaluated once.

\textbf{\textit{k}-fold Cross-Validation}. In the case of limited data, however, one would strive to involve as much data in training as possible. \textit{k}-fold Cross-Validation~\cite{cite:IJCAI95CV} splits the given data into \textit{k} folds, runs the algorithm using $\frac{k - 1}{k}$ data, validates it using the rest $\frac{1}{k}$ data and then repeats \textit{k} times, averaging the results. \textit{k}-fold Cross-Validation exploits all the available data at the cost of \textit{k} runs. \textbf{Leave-one-out} is one particular case of \textit{k}-fold Cross-Validation with \textit{k} equaled to the number of training data. It can be applied in the extreme circumstance where the labeled data is particularly scarce.

Although learning in a single domain is well defined, the assumption that the test distribution is the same as the training distribution is often violated in real-world applications. For example, clinical data is collected from patients while the algorithm developed from the data would be tested among ordinary people. Models learned from patients cannot apply to normal people because of the dataset bias. Thus it would be of great significance if models can learn from different domains. Still, it is difficult to formalize learning from two arbitrary domains. A well-defined paradigm is to learn from two domains under the \textit{covariate shift} assumption~\cite{blitzer_learning_2008,cite:ECCV10Office,cite:ML10LT}. 

\subsection{Model Selection in Domain Adaptation}

\textbf{Importance-Weighted Cross-Validation}. When encountering covariate shift problems, the model selection methods for learning in a single domain would fail to identify the best model for the target domain. The estimation in both the \textit{training set} and the \textit{validation set} would be biased and cannot reflect the test risk. \citet{cite:JMLR07IWCV} proposed Importance-Weighted Cross-Validation (IWCV) to perform cross-validation under covariate shift with the aid of known density ratio. The validation risk is weighted to be an unbiased estimator of the target risk. Though unbiased, such an estimator has unbounded variance. Moreover, when the density ratio is not readily available, IWCV needs to estimate the density ratio by a single multi-dimensional normal distribution, which is cumbersome and inaccurate.

Later, \citet{cite:NIPS10LB} revealed that the variance of the importance-weighted methods can be bounded by a family of R\'{e}nyi divergence~\cite{cite:RENYI}. While the variance is bounded by the R\'{e}nyi divergence, neither the variance itself nor the bound of the variance is lowered.

\textbf{Transfer Cross-Validation}. By considering both marginal and conditional distributions in different domains, ~\citet{cite:TrCV} proposed Transfer Cross-Validation (TrCV) for model selection under both marginal and conditional shifts. To approximate the conditional distribution, TrCV requires labeled target data to assist the model selection process. TrCV incurs similar controversy as tuning hyperparameters by Target Risk. In Deep UDA, one would prefer a model selection method that works without target labeled data but still correlates well to Target Risk with statistical guarantees.

The mentioned model selection methods both originated in the era before deep learning. They both work in the \textit{covariate} level, i.e., they validate models directly based on the input data. In the age of deep learning, it is natural to extend these models to work in the \textit{feature} level. While intuitive, the theoretical property of such an ``embedded'' validation remains unknown. We are the first to validate models on the feature level and provide the theoretical insight behind the embedded validation. The validation yields a tighter bound on its variance and remains unbiased. The variance can be further lowered by employing a control variate method.

\section{Preliminaries}

%In this section, we provide the preliminary math knowledge such that this paper is self-contained. 

\subsection{R\'{e}nyi Divergence}

We first introduce the notation of R\'{e}nyi divergence defined in~\cite{cite:RENYI,cite:NIPS10LB}. R\'{e}nyi divergence between distributions $p$ and $q$ is defined as $D _ { \alpha } ( p \| q ) = \frac { 1 } { \alpha - 1 } \log _ { 2 } \sum _ { x } p ( x ) \left( \frac { p ( x ) } { q ( x ) } \right) ^ { \alpha - 1 }$, where hyperparameter $\alpha \ge 0$ and $\alpha \neq 1$. Note that $\underset{\alpha \rightarrow 1}{\lim} D_\alpha(p \| q) = KL(p \| q)$, which is the widely-used Kullback-Leibler divergence. R\'{e}nyi divergence satisfies the properties of a well-defined divergence: it is non-negative and $D _ { \alpha } ( p \| q ) = 0$ if and only if $p = q$. For brevity, another notation of R\'{e}nyi divergence is adopted:
\begin{equation}
d _ { \alpha } ( p \| q ) = 2 ^ { D _ { \alpha } ( p \| q ) } = \left[ \sum _ { x } \frac { p ^ { \alpha } ( x ) } { q ^ { \alpha - 1 } ( x ) } \right] ^ { \frac { 1 } { \alpha - 1 } }.
\end{equation}

\subsection{Control Variates}

\label{sec:control_variate}

The \textit{control variates} method~\cite{lemieux_control_2017}, widely used in Monte Carlo methods, is an effective technique to reduce variance. Suppose the statistic $z$ is an unbiased estimator of an unknown parameter $\zeta$, i.e. $\mathbb{E}[z] = \zeta $. However, an unbiased estimator would never be accurate if its variance Var$[z]$ is high. To reduce its variance, we can find another related unbiased estimator $t$ such that $\mathbb{E}[t] = \tau$, where $\tau$ is the parameter that $t$ tries to estimate. Then we can construct a new estimator parameterized by a constant $\eta$,
\begin{equation}
\label{eq:control_variate}
z^\star = z + \eta (t - \tau).
\end{equation}
It is straightforward to show that $z^\star$ is still unbiased thanks to the linear property of the expectation operation:
\begin{equation*}
\mathbb{E}[z^\star]= \mathbb{E}[z] + \eta \mathbb{E}[t - \tau]= \zeta + \eta (\mathbb{E}[t] - \mathbb{E}[\tau]) = \zeta.
\end{equation*}
The variance of $z^\star$ can be computed as
\begin{equation}
\begin{aligned}
\mathrm{Var}[z^\star] &= \mathrm{Var}[z + \eta (t - \tau)] \\
&= \eta^2 \mathrm{Var}[t]  + 2\eta\mathrm{Cov}(z, t) + \mathrm{Var}[z] ,\\
\end{aligned}
\end{equation}
which is a quadratic form of $\eta$ and have a global optimum
\begin{equation}
\min \mathrm{Var}[z^\star] = (1 - \rho_{z,t}^2) \mathrm{Var}[z], \text{ when} \; 
\hat{\eta} = - \frac{\mathrm{Cov}(z, t)}{\mathrm{Var}[t]}, \\
\end{equation}
where $\rho_{z,t}$ is the correlation coefficient of $z$ and $t$. Since $0 \le |\rho_{z,t}| \le 1$, the variance is reduced: $\mathrm{Var}[z^\star] \le \mathrm{Var}[z]$. 

In essence, the control variate method finds a correlated and unbiased variable and subtracts it with a proper coefficient, thus making the estimator deviate less from the expectation.

\section{Method}

%In this section, we first formalize the model selection problem under the covariate shift assumption and then propose our solution.
%
%\subsection{Problem Setup}

In Deep UDA, we learn towards a joint distribution $J(\mathbf{x}, d)$, where $\mathbf{x}$ is the input associated with the label $y$, and $d$ is a Bernoulli variable indicating the domain that $\mathbf{x}$ belongs to. Let $p(\mathbf{x}) = J(\mathbf{x}, d | d = 1)$ denote the \textit{source domain} distribution and $q(\mathbf{x}) = J(\mathbf{x}, d | d = 0)$ denote the \textit{target domain} distribution. The domain shift $p(\mathbf{x}) \neq q(\mathbf{x})$ presents a major challenge for domain adaptation. Meanwhile, \textit{covariate shift} is a common assumption that says $p(y|\mathbf{x}) = q(y|\mathbf{x})$, i.e. the class label of the input is independent of the domain. Samples drawn \textit{i.i.d} from the joint distribution $J(\mathbf{x}, d)$ form our dataset: \textit{source domain} observations $\mathcal { D } _ { s } = \left\{ \left( \mathbf { x } _ { i } ^ { s } , y _ { i } ^ { s } \right) \right\} _ { i = 1 } ^ { n _ { s } }$ with $d = 1$, \textit{target domain} observations $\mathcal { D } _ { t } = \left\{ \left( \mathbf { x } _ { i } ^ { t } , y _ { i } ^ { t } \right) \right\} _ { i = 1 } ^ { n _ { t } }$ with $d = 0$ ($y _ { i } ^ { t }$ is not accessible in the training phase).

Model selection is to find $\hat{g} = {\arg \min}_{g \in G} \mathbb{E}_{\mathbf{x} \sim q}\ell(g(\mathbf{x}), y)$, where $G$ is the model space where each model maps the input $\mathbf{x}$ to the output $\hat{y}$, and $\ell(\cdot, \cdot)$ is the loss function. In reality, though, it is infeasible to search through the whole model space, and we opt to find $\hat{g} = {\arg \min}_{g \in G_m} \mathbb{E}_{\mathbf{x} \sim q}\ell(g(\mathbf{x}), y)$, where $G_m = \left\{ g_i \right\}_{i=1}^{m}$ is a finite set of candidate models. 

The difficulty for model selection in Deep UDA arises from the fact that $y$ is inaccessible when $\mathbf{x} \sim q$. Hence, we have to estimate $\mathbb{E}_{\mathbf{x} \sim q}\ell(g(\mathbf{x}), y)$ with the help of labeled source data $\mathbf{x}\sim p$.
Considering that deep learning models usually learn a discriminative feature representation and then perform downstream tasks, we split $g$ into two functions: $g(\mathbf{x}) = \mathcal{T}({\bm f})$, where ${\bm f} = F(\mathbf{x})$. Here $F$ is the feature extractor and $\mathcal{T}$ takes the feature ${\bm f}$ to perform specific tasks.

\subsection{Importance-Weighted Cross-Validation}

The main challenge for model selection in Deep UDA is that the target risk $\mathcal{R}(g) = \mathbb{E}_{\mathbf{x} \sim q}\ell(g(\mathbf{x}), y)$ is defined over the target domain distribution $q$ without any labeled data. If density ratio (a.k.a. importance weights) $w(\mathbf{x}) = \frac{q(\mathbf{x})}{p(\mathbf{x})}$ is known, following~\citet{cite:JMLR07IWCV}, we can obtain
\begin{equation}
\label{eq:IWCV_unbias}
\begin{aligned}
\mathbb{E}_{\mathbf{x} \sim p} w(\mathbf{x}) \ell(g(\mathbf{x}), y) &= \mathbb{E}_{\mathbf{x} \sim p} \frac{q(\mathbf{x})}{p(\mathbf{x})} \ell(g(\mathbf{x}), y) \\
&= \int_p \frac{q(\mathbf{x})}{p(\mathbf{x})} \ell(g(\mathbf{x}), y) p(\mathbf{x}) \rm{d}\mathbf{x} \\
&= \int_q \ell(g(\mathbf{x}), y) q(\mathbf{x}) \rm{d}\mathbf{x} \\
&= \mathbb{E}_{\mathbf{x} \sim q}\ell(g(\mathbf{x}), y) \\
&= \mathcal{R}(g) ,
\end{aligned}
\end{equation} which means $\frac{1}{n_s} \Sigma_{i=1}^{n_s} (w(\mathbf{x}_i^s) \ell(g(\mathbf{x}_i^s), y_i^s))$ is an \emph{unbiased} estimator of the target risk $\mathcal{R}(g)$.

For brevity, we denote $w(\mathbf{x}) \ell(g(\mathbf{x}), y) $ by $\ell_w$. As shown in ~\citet{cite:NIPS10LB} (Lemma 2), the variance of importance-weighted cross-validation is bounded by R\'{e}nyi divergence:
\begin{equation}
\label{eq:variance_bound}
\begin{aligned}
\mathrm{Var}_{\mathbf{x} \sim p}[\ell_w] &= \mathbb{E}_{\mathbf{x} \sim p}[(\ell_w) ^2] - (\mathbb{E}_{\mathbf{x} \sim p}[\ell_w] ) ^2 \\
& \le d _ { \alpha + 1 } ( q \| p ) \mathcal{R} ( g ) ^ { 1 - \frac { 1 } { \alpha } } - \mathcal{R} ( g ) ^2.
\end{aligned}
\end{equation}

\subsection{Deep Embedded Validation}

\label{sec:dev}

As shown in Eq.~\eqref{eq:variance_bound}, the variance of importance-weighted cross-validation (IWCV) is bounded by R\'{e}nyi divergence between distributions $ p$ and $q$. However, neither the variance of IWCV nor its bound is lowered as domain adaptation goes on, given that $p$ and $q$ stay still.
Recent work \cite{cite:NIPS18CDAN} (Figure 2(c)) shows that the distribution divergence becomes smaller after feature adaptation. While $p$ and $q$ stay still in the input space, better adaptation model tends to show lower distribution divergence in the feature space. Note that these models can only reduce the distribution divergence rather than closing it, implying that it is still necessary to develop a validation method for Deep UDA.

These observations inspire us to step from the \textit{covariate} space to the \textit{feature} space. Let $p_{{\bm f}} $ and $ q_{{\bm f}}$ be the feature distributions of the source domain and the target domain respectively. Deep domain adaptation models usually close the domain gap by learning domain-invariant features, which implies $d_{\alpha + 1} (q_{{\bm f}} \| p_{{\bm f}})$ is generally smaller than $d_{\alpha + 1} (q \| p)$. Thereby, we propose to embed the learned deep features into the validation procedure, resulting in an embedded density ratio estimation $w_{\bm f}(\mathbf{x}) = \frac{q_{\bm f}(\mathbf{x})}{p_{\bm f}(\mathbf{x})}$. By changing from the \textit{covariate} level to the \textit{feature} level, we can also conclude that $\frac{1}{n_s} \Sigma_{i=1}^{i=n_s} (w_{{\bm f}}(\mathbf{x}_i^s) \ell(g(\mathbf{x}_i^s), y_i^s))$ is an unbiased estimator of the target risk, with its variance $\mathrm{Var}_{\mathbf{x} \sim p_{\bm f}}[\ell_{w_{\bm f}}]$ bounded by $ d _ { \alpha + 1 } ( q_{{\bm f}} \| p_{{\bm f}} ) \mathcal{R} ( g ) ^ { 1 - \frac { 1 } { \alpha } } - \mathcal{R} ( g ) ^2$, which is generally smaller than $ d _ { \alpha + 1 } ( q \| p ) \mathcal{R} ( g ) ^ { 1 - \frac { 1 } { \alpha } } - \mathcal{R} ( g ) ^2$.

As can be verified in Eq.~\eqref{eq:IWCV_unbias}, IWCV requires an important assumption that the support of $p$ contains the support of $q$, i.e. $\mathbf{supp}(p) \supset \mathbf{supp}(q)$, where $ \mathbf{supp}(p) = \left\{ {\bf x} | p({\bf x}) \neq 0 \right\} $.  
If the assumption is violated, the importance weights can grow to infinity. Before aligning distributions $p$ and $q$, it is highly possible that the assumption is violated, especially image and text data with high-dimensional input covariates. After feature learning and adaptation, the deep features are made more compact and domain-invariant. The assumption on the support of $q$ in $p$ can hold well in the learned feature space.

\subsection{Discriminative Density Ratio Estimation}
\label{sec:density_ratio_estimation}
Density ratio is not readily accessible in pragmatic applications. Here we adopt an approach similar in~\cite{bickel_discriminative_2007}, using Bayesian formula to derive density ratio from a model that discriminates between source and target samples:
\begin{equation}
\label{eq:density_ratio}
\begin{aligned}
w_{\bm f}(\mathbf{x}) &= \frac{q_{\bm f}(\mathbf{x})}{p_{\bm f}(\mathbf{x})} = \frac{J_{{\bm f}}(\mathbf{x} | d=0)}{J_{{\bm f}}(\mathbf{x} | d=1)} \\
&= \frac { J_{\bm f} ( d = 1 ) } { J_{\bm f} ( d = 0 ) } \frac { J_{\bm f} ( \mathbf { x } ) J_{\bm f} ( d = 0 | \mathbf { x } ) } { J_{\bm f} ( \mathbf { x } ) J_{\bm f}( d = 1 | \mathbf { x } ) } \\ 
&= \frac { J_{\bm f} ( d = 1 ) } { J_{\bm f} ( d = 0 ) } \frac { J_{\bm f} ( d = 0 | \mathbf { x } ) } { J_{\bm f} ( d = 1 | \mathbf { x } ) } \\
&= \frac{n_s}{n_t} \frac { J_{\bm f} ( d = 0 | \mathbf { x } ) } { J_{\bm f} ( d = 1 | \mathbf { x } ) },
\end{aligned}
\end{equation}
where $J_{\bm f}$ is the joint distribution in the feature space. 

As can be seen in Eq.~\eqref{eq:density_ratio}, density ratio can be decomposed into two parts: $\frac { J_{\bm f} ( d = 0 | \mathbf { x } ) } { J_{\bm f} ( d = 1 | \mathbf { x } ) }$ and a constant factor $\frac { J_{\bm f} ( d = 1 ) } { J_{\bm f} ( d = 0 ) }$. The former can be estimated by a discriminative model to distinguish source examples from target examples. The latter constant factor does not vary with models and can be estimated with the sample sizes of both domains. Note that the discriminative model is trained on the readily-available domain information $d$, which follows a fully supervised learning scheme. A two-layer logistic regression model will be accurate enough for estimating the density ratio.

\subsection{Variance Reduction by Control Variate}

In Section~\ref{sec:dev}, it is clear that using embedded deep feature representations for model selection will benefit from a lower bound on the variance of the target risk estimation. Meanwhile, we can explicitly reduce the variance by adopting the \textit{control variate} method described in Section~\ref{sec:control_variate}.

To reduce the variance $\mathrm{Var}_{\mathbf{x} \sim p_{\bm f}}[\ell_{w_{{\bm f}}}]$ of target risk estimate, there are two candidates for the control variate: $w_{{\bm f}}$ and $\ell$. However, the expectation of $\ell$ is not only unknown but also task-specific, depending on the choice of $\ell(\cdot, \cdot)$. By contrast, the expectation of $w_{{\bm f}}$ remains independent of the model $\mathcal{T}$,
\begin{equation}
\begin{aligned}
\mathbb{E}_{\mathbf{x} \sim p_{{\bm f}}} w_{{\bm f}}(\mathbf{x}) &= \mathbb{E}_{\mathbf{x} \sim p_{{\bm f}}} \frac{q_{\bm f}(\mathbf{x})}{p_{\bm f}(\mathbf{x})} \\
&= \int \frac{q_{\bm f}(\mathbf{x})}{p_{\bm f}(\mathbf{x})} p_{\bm f}(\mathbf{x}) \rm{d}\mathbf{x} = 1.
\end{aligned}
\end{equation}

By plugging $w_{{\bm f}}$ into Eq.~\eqref{eq:control_variate} as a control variate for the target risk estimation, we can embed features ${\bm f}$ in the estimator as
\begin{equation}
\begin{aligned}
\mathcal{R}_\text{DEV} &= \frac{1}{n_s} \sum_{i=1}^{n_s} w_{\bm f}(\mathbf{x}_i^s) \ell(g(\mathbf{x}_i^s), y_i^s) \\ 
&+ \frac{\eta}{n_s} \sum_{i=1}^{n_s} \: [w_{\bm f}(\mathbf{x}_i^s) - \mathbb{E}_{\mathbf{x} \sim p_{{\bm f}}} w_{{\bm f}}(\mathbf{x}) ] \\
&=\frac{1}{n_s} \sum_{i=1}^{n_s} \: [ \ell_{w_{\bm{f}}} + \eta (w_{\bm f}(\mathbf{x}_i^s) - 1)],
\end{aligned}
\end{equation}
where $\eta$ is the optimal coefficient estimated by 
\begin{equation}
\eta = - \frac{\widehat{\mathrm{Cov}}(\ell_{w_{{\bm f}}}, w_{{\bm f}})}{\widehat{\mathrm{Var}}[w_{{\bm f}}]}.
\end{equation}

The complete validation procedure, which is called \textit{Deep Embedded Validation} (\textbf{DEV}), is described in Algorithms~\ref{alg:get_risk} and \ref{alg:dev}. DEV is tailored to Deep UDA models by embedding adapted deep feature representations into model selection. It is an unbiased estimation of the target risk while its variance is bounded by a theoretical guarantee and is further reduced by the control variate method. Note that the control variate method can be applied not only on deep models but also on shallow models. But deep models benefit more by having smaller R\'{e}nyi divergence, thus having a lower upper bound.

\begin{algorithm}[htb]
	\caption{\centering GetRisk}
	\label{alg:get_risk}
	\begin{algorithmic}
		\STATE {\bfseries Input:} Candidate model $g(\mathbf{x}) = \mathcal{T}(F(\mathbf{x}))$ \\	
		\quad  \quad \;\;\; {Training set} $\mathcal { D } _ \text{tr} = \left\{ \left( \mathbf { x } _ { i } ^ \text{tr} , y _ { i } ^ \text{tr} \right) \right\} _ { i = 1 } ^ { n _ \text{tr} }$  \\ 
		\quad  \quad \;\;\;  {Validation set} $\mathcal { D } _ \text{v} = \left\{ \left( \mathbf { x } _ { i } ^ \text{v} , y _ { i } ^ \text{v} \right) \right\} _ { i = 1 } ^ { n _ \text{v} }$  \\ 
		\quad  \quad \;\;\;  {Test set}  $\mathcal { D } _ \text{ts} = \left\{ \left( \mathbf { x } _ { i } ^ \text{ts}  \right) \right\} _ { i = 1 } ^ { n _ \text{ts} }$ \\
		\quad  \quad \;\;\;  $\mathcal{D}_s$ is partitioned into $\mathcal{D}_\text{tr}$ and $\mathcal{D}_
		\text{v}$\\
		\STATE {\bfseries Output:} DEV Risk $\mathcal{R}_{\text{DEV}}(g)$ of model $g$
		
		\vspace{5pt}
		
		\STATE { Compute} features and predictions using model $g$:
		
		\quad \quad \quad \quad $\mathcal{F}_\text{tr} = \left\{ {\bm f}_i^\text{tr} \right\}_{i=1}^{n_\text{tr}}$, $\mathcal{F}_\text{ts} = \left\{ {\bm f}_i^
		\text{ts} \right\}_{i=1}^{n_\text{ts}}$

		\quad \quad \quad \quad $\mathcal{F}_\text{v}= \left\{ {\bm f}_i^\text{v} \right\}_{i=1}^{n_\text{v}}$, $\mathcal{Y}_\text{v} = \left\{ \hat{y}_i^\text{v} \right\}_{i=1}^{n_\text{v}}$ \\
		
		Train a two-layer logistic regression model $M$ to classify $\mathcal{F}_\text{tr}$ and $\mathcal{F}_\text{ts}$ (label $\mathcal{F}_\text{tr}$ as 1 and $\mathcal{F}_\text{ts}$ as 0) \\
		
		Compute $w_{{\bm f}}(\mathbf{x}_i^\text{v}) = \frac{n_\text{tr}}{n_\text{ts}} \frac{1 - M({\bm f}_i^\text{v})}{M({\bm f}_i^\text{v})}$, $W = \left\{  w_{{\bm f}}(\mathbf{x}_i^\text{v}) \right\}_{i=1}^{n_\text{v}}$ \\
		
		Compute weighted loss $L = \left\{ w_{{\bm f}}(\mathbf{x}_i^\text{v}) \ell(\hat{y}_i^\text{v}, {y}_i^\text{v}) \right\}_{i=1}^{n_\text{v}}$ \\
		
		Estimate coefficient $\eta = - \frac{\widehat{\mathrm{Cov}}(L, W)}{\widehat{\mathrm{Var}}[W]}$ \\
		
		Compute DEV Risk:
		
		\quad \quad \quad \quad $\mathcal{R}_{\text{DEV}}(g)= \mathrm{mean}(L) + \eta \mathrm{mean}(W) - \eta$ \\
	\end{algorithmic}
\end{algorithm}

\begin{algorithm}[htb]
	\caption{\centering Deep Embedded Validation (DEV)}
	\label{alg:dev}
	\begin{algorithmic}
		\STATE {\bfseries Input:} A set of candidate models $G_m = \left\{ g_i(\mathbf{x})  \right\}_{i=1}^{m} $ \\
		
		\STATE {\bfseries Output:} The best model $(G_m)_{\hat{i}}$
		
		\vspace{5pt}
		
		\STATE Get DEV Risks of all models
		$\mathcal{R} = \{\text{GetRisk}(g_i)\}_{i=1}^m$ \\
		
		Rank the best model $\hat{i} = {\arg \min}_{1 \le i \le m} \mathcal{R}_i$ \\
	\end{algorithmic}
\end{algorithm}

\subsection{Beyond Feature Adaptation Methods}

The above discussion focuses on feature adaptation models for Deep UDA. These models learn features to reduce the distribution shift, which also reduces the variance of target risk estimation. However, DEV is not limited to the feature adaptation models. For example, generative models that generate an auxiliary domain \cite{cite:CVPR18IIT} can be formalized as generating another distribution $\hat{p}$ that is closer to $q$. In such a scenario, samples generated by $\hat{p}$ naturally have labels, and DEV can be carried out between $\hat{p}$ and $q$. 

\begin{figure*}[htb]
	\centering
	\subfigure[]{
		\label{fig:input_density}
		\includegraphics[width=0.23\textwidth]{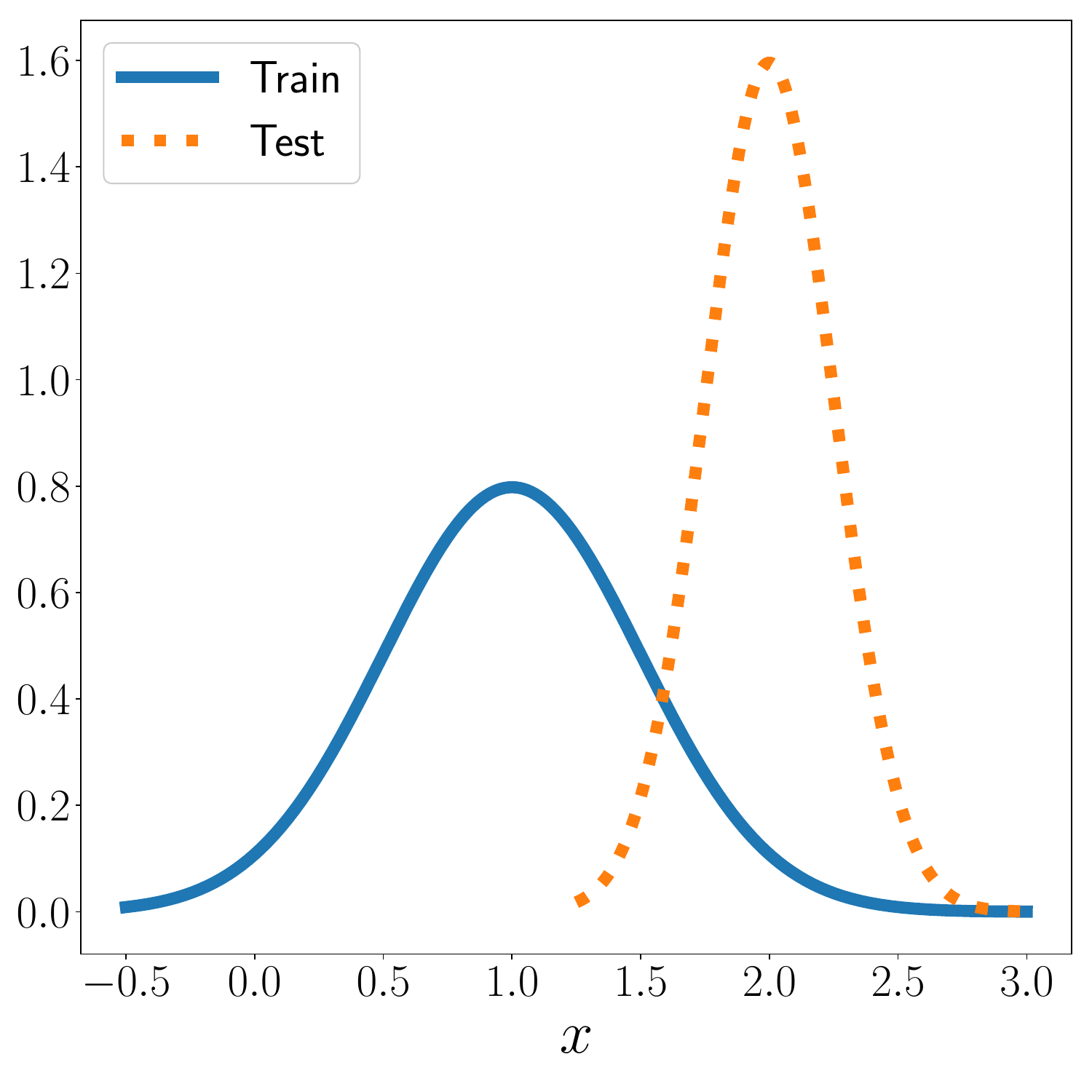}}
		\hfil
	\centering
	\subfigure[]{
		\label{fig:dataset}
		\includegraphics[width=0.23\textwidth]{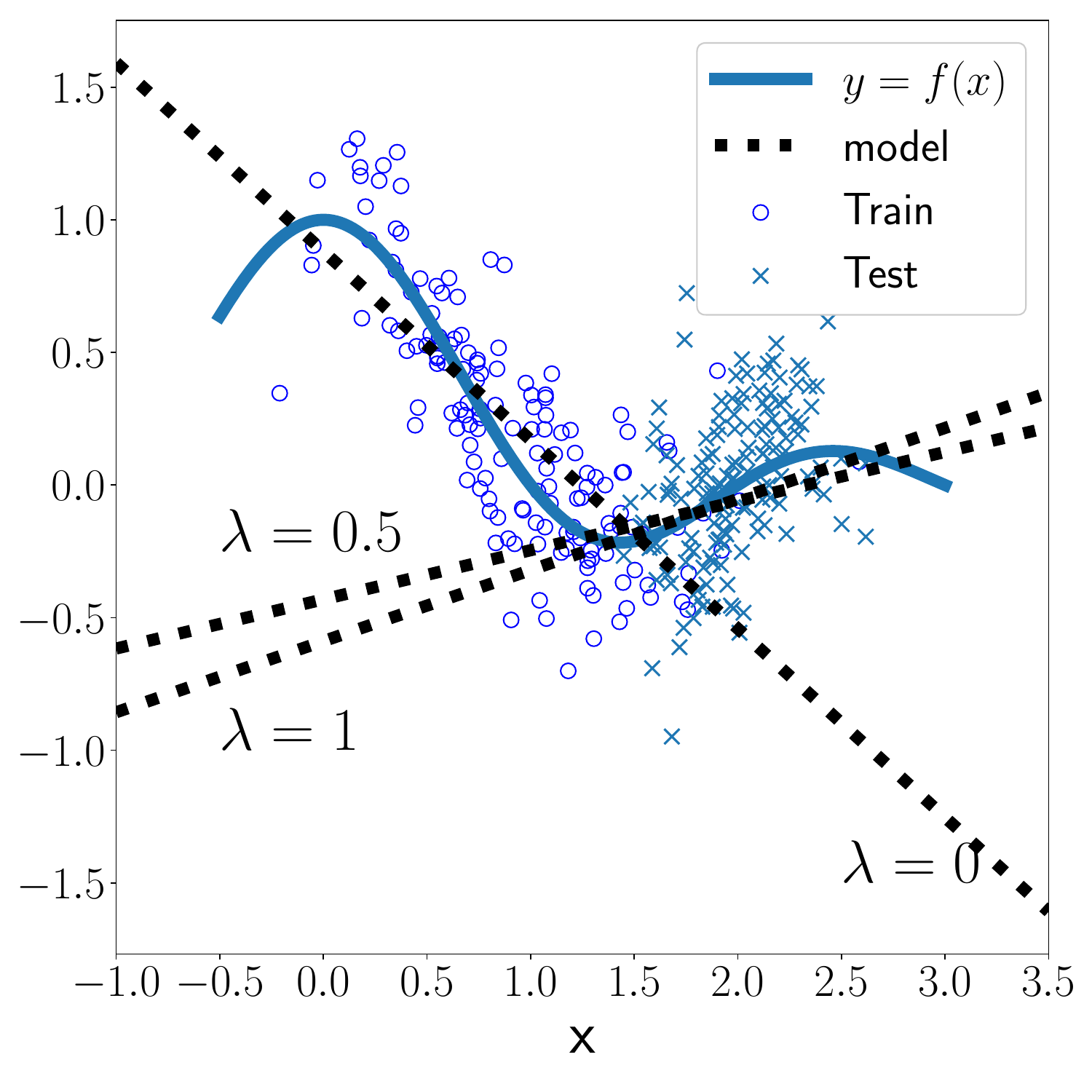}}
		\hfil
	\centering
\subfigure[]{
	\label{fig:means_and_stds}
	\includegraphics[width=0.23\textwidth]{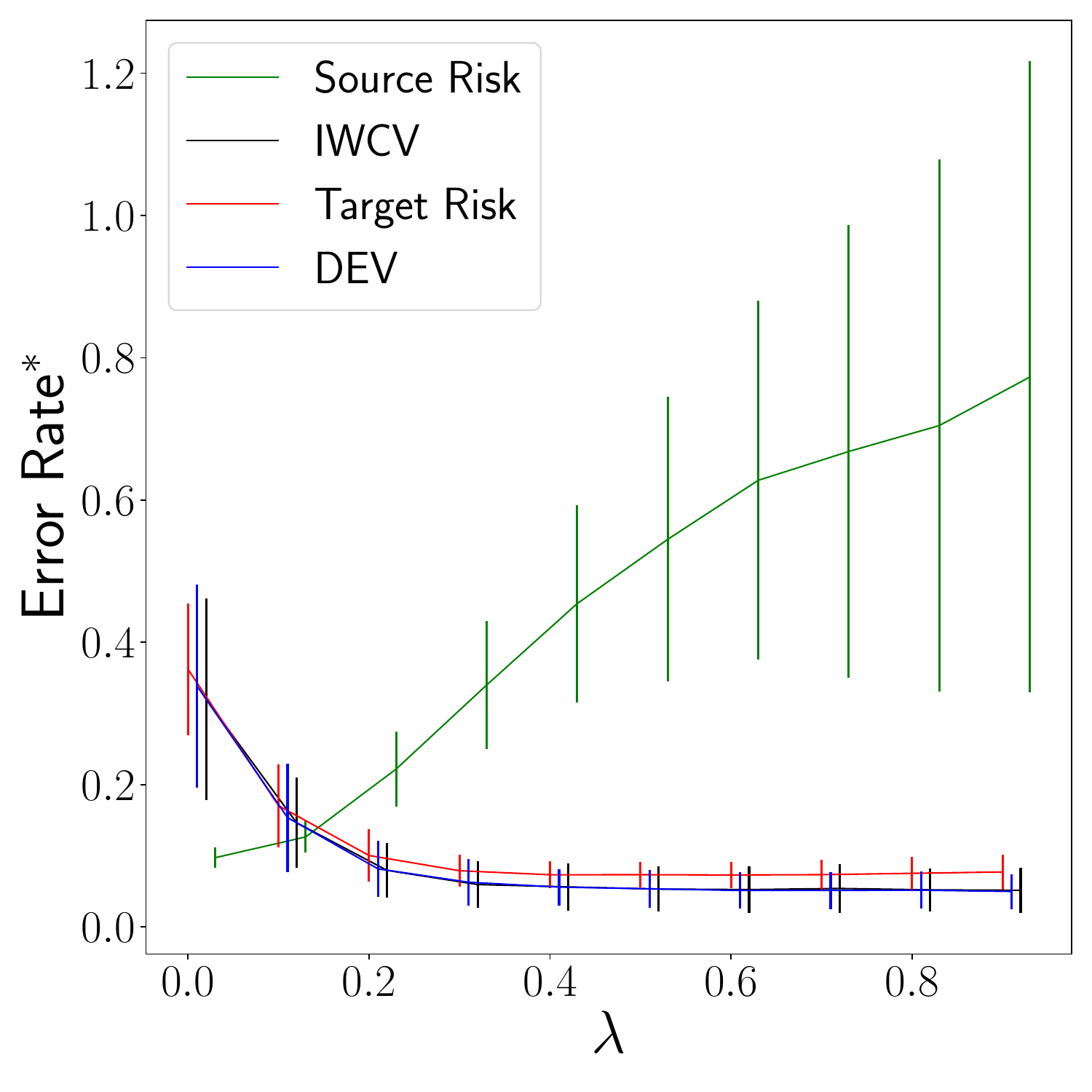}}
		\hfil
	\centering
\subfigure[]{
	\label{fig:stds_in_detail}
	\includegraphics[width=0.23\textwidth]{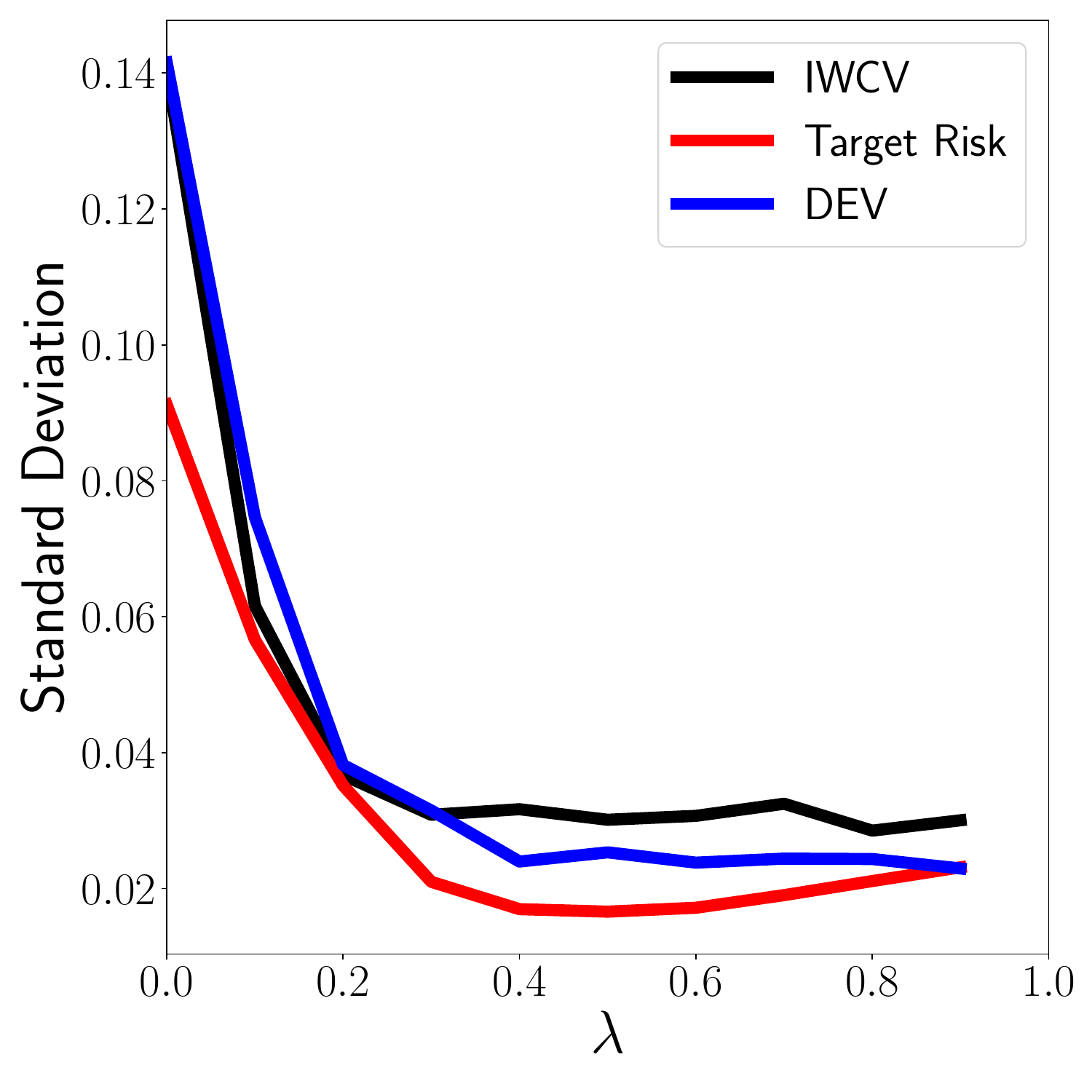}}
	\vskip -0.1in
	\caption{An example of model selection in a toy dataset for regression. (a) Input data density with covariate shift. (b) Dataset and the underlying objective function as well as several candidate models with different hyperparameters $\lambda$. (c) Comparison among different validation methods. (d) The standard deviation in details. (*) The error rate is used to select models. It is not the final reported accuracy.}
\end{figure*}

\section{Experiments}

In this section, we conduct a series of experiments to empirically evaluate the proposed {DEV} approach. We first play around with a toy dataset and then dive into Deep UDA models. With deep models, we try the following learning rates: $\left\{ 10^{-2}, 10^{-2.5},10^{-3},10^{-3.5},10^{-4} \right\} $. Other hyperparameters are specified in their subsections respectively. Model selection is conducted based on different methods: (\textbf{1}) using source error to select models (\textbf{Source Risk}); (\textbf{2}) using target error to select models (\textbf{Target Risk}); (\textbf{3}) \textbf{IWCV}~\cite{cite:JMLR07IWCV}, with importance weights generated based on pre-trained feature representations; (\textbf{4}) \textbf{DEV}. The code of DEV is available at \url{https://github.com/thuml/Deep-Embedded-Validation}. 

There are several clarifications. (\textbf{1}) \citet{cite:JMLR07IWCV} fits a multi-variate Gaussian to approximate the density ratio. Gaussians always underfit deep features and yield extremely unstable weights. Thus we give it the benefit to estimate the density ratio by using pre-trained deep features to train a binary classifier as described in Section~\ref{sec:density_ratio_estimation}. (\textbf{2}) In Section~\ref{sec:Digits}, there is no standard model for feature extraction on Digits dataset, so IWCV is not reported. For other datasets, the features used by IWCV are extracted by ResNet-50 \cite{cite:CVPR16DRL}. (\textbf{3}) For a fair comparison, we also list the results reported in their original papers (denoted as Original).

As mentioned in Section~\ref{sec:validation}, cross-validation is needed when labeled data is scarce. In Deep UDA, while we do not have labeled data in the target domain, a large number of labeled data in the source domain are available. Since DEV is carried out on the source data, we can split the source data into train/validation set before learning. That said, we use the hold-out validation method throughout all experiments.

\begin{table*}[htbp]
	\centering
	\addtolength{\tabcolsep}{-3pt} 
	\caption{Accuracy (\%) of \textbf{MCD} \cite{cite:CVPR18MCD} by different validation methods on {VisDA} dataset.}
	\label{tab:MCD}%
	\vskip 0.05in
	\begin{center}
	\resizebox{\textwidth}{!}{%
		\begin{tabular}{lccccccccccccc}
			\toprule
			Method & plane & bcycl & bus & car & horse & knife & mcycl & person & plant& sktbrd & train & truck & mean \\
			\midrule
			Original~\cite{cite:CVPR18MCD} & 87.00  & 60.90  & 83.70  & 64.00  & 88.90  & 79.60  & 84.70  & 76.90  & 88.60  & 40.30  & 83.00  & 25.80  & 71.90  \\
			{Source Risk} & 84.39  & 54.11  & 69.15  & 46.37  & 80.49  & 80.45  & 85.04  & 65.24  & 87.22  & 36.86 & 78.04  & 28.91 & 66.36  \\
			{IWCV} & 80.80  & 48.26  & 68.89 & 69.89  & 59.62  & 37.49  & 46.51  & 23.83  & 75.25  & 46.56  & 83.43 & 16.08 & 54.72  \\
			\textbf{DEV} (w/o control variate) & 84.21  &	63.95 	& 79.00 &	59.53 &	85.83 &	75.06 	& 87.05 &	71.65 &	89.85 &	47.37 &	72.29 &	26.81 &	70.22 \\
			\textbf{DEV} & 81.83  & 53.48  & 82.95  & 71.62  & 89.16  & 72.03  & 89.36  & 75.73  & 97.02  & 55.48 & 71.19  & 29.17  & 72.42   \\
			\midrule
			{Target Risk} (Upper Bound) & 81.95  & 53.60  & 83.07  & 72.02  & 89.25  & 72.15  & 89.55  & 75.83  & 97.10  & 55.57 & 71.19  & 29.27  & 72.55  \\
			\bottomrule
		\end{tabular}%
	}
	\end{center}
\end{table*}%

\begin{table*}[!htbp]
	\centering
	\caption{Accuracy (\%) of \textbf{CDAN} \cite{cite:NIPS18CDAN} by different validation methods on {Office-31} dataset.}
	\label{tab:cdan}
	\vskip 0.05in
	\begin{tabular}{lccccccc}
		\toprule
		Method & A $\rightarrow$ W & D $\rightarrow$ W & W $\rightarrow$ D & A $\rightarrow$ D & D $\rightarrow$ A & W $\rightarrow$ A & Avg \\
		\midrule
		Original~\cite{cite:NIPS18CDAN} & {93.10} & 98.60& {100.00} & {92.90} & {71.00} & {69.30} & {87.50} \\
		{Source Risk} & {85.95} & {98.60}& {100.00} & {84.59} & {65.00} & {61.34} & {82.58} \\
		{IWCV} & {88.95} & {95.30} & {97.00} & {87.59}& {65.37} & {67.95} & {83.69} \\
		\textbf{DEV} & {93.23} & {98.40} & {100.00} & {92.81} & {70.89} & {71.15} & {87.75} \\
		\midrule
		{Target Risk} (Upper Bound) & {93.33} & {100.00} & {100.00} & {93.06}& {71.10} & {71.45} & {88.16} \\
		\bottomrule
	\end{tabular}
\end{table*}

\begin{table*}[hbt]
	\centering 
	\caption{Accuracy (\%) of \textbf{GTA} \cite{cite:CVPR18GenerateAdapt} by different validation methods on {Digits} dataset.}
	\label{tab:digit}
	\vskip 0.05in
		\begin{tabular}{lcccc}
	\toprule
	Method & USPS $\rightarrow$ MNIST & MNIST $\rightarrow$ USPS & SVHN $\rightarrow$ MNIST & Avg \\
	\midrule
	Original~\cite{cite:CVPR18GenerateAdapt} & 95.30 & 90.80 & 92.40 & 92.83  \\
	{Source Risk }& 92.03 & 85.92 & 77.58 & 85.18 \\
	\textbf{DEV} & 96.93 & 92.54 & 93.18 & 94.22 \\
	\midrule
	{Target Risk} (Upper Bound) & 97.03 & 92.97 & 93.51 &94.50 \\
	\bottomrule
\end{tabular}
\end{table*}

\begin{table*}[!htbp]
	\centering
	\addtolength{\tabcolsep}{-3pt} 
	\caption{Accuracy (\%) of \textbf{PADA} \cite{cite:ECCV18PADA} by different validation methods on {Office-31} dataset.}
	\label{tab:PADA}
	\vskip 0.05in
	\resizebox{\textwidth}{!}{%
	\begin{tabular}{lccccccc}
		\toprule
		Method & \begin{small}
			A31 $\rightarrow$ W10
		\end{small} & \begin{small}
			D31 $\rightarrow$ W10
		\end{small} & \begin{small}
			W31 $\rightarrow$ D10
		\end{small} & \begin{small}
			A31 $\rightarrow$ D10
		\end{small} & \begin{small}
			D31 $\rightarrow$ A10
		\end{small} & \begin{small}
			W31 $\rightarrow$ A10
		\end{small} & Avg \\
		\midrule
		Original~\cite{cite:ECCV18PADA} & 86.54 & 99.32 & 100.00 & 82.17 & 92.69 & 95.41 & 92.69 \\
		{Source Risk }& 70.17 & 98.30 & 99.32 & 76.17 & 88.51 & 90.92 & 87.23 \\
		{IWCV} & 82.38 & 97.00 & 96.42 & 78.96 & 89.16 & 92.23 & 89.36 \\
		\textbf{DEV} & 87.80 & 100.00 & 100.00 & 82.94 & 92.84 & 95.23 & 93.15 \\
		\midrule
		{Target Risk} (Upper Bound) & 87.80 & 100.00 & 100.00 & 83.59 & 93.00 & 95.66 & 93.34 \\
		\bottomrule
	\end{tabular}
	}
\end{table*}

\subsection{Toy Dataset}

Figures~\ref{fig:input_density} and \ref{fig:dataset} show a toy regression data following the protocol of~\citet{cite:JMLR07IWCV}. Data points lie on $y=\frac{\sin(\pi x)}{\pi x}$ with random noise sampled from normal distribution $\mathcal{N}(0, (\frac{1}{4})^2)$. The marginal distribution of $x$ differs in the training set and test set (Figure~\ref{fig:input_density}), explaining the \textit{covariate shift} problem. Here $p(x) = \mathcal{N}(x|1, (\frac{1}{2})^2), q(x) = \mathcal{N}(x|2, (\frac{1}{4})^2)$. The density ratio $w(x)$ can be computed analytically. Candidate models for the toy problem are AIWLS models~\cite{cite:JMLR07IWCV} with different hyperparameters $\lambda$ ranging from $0.0$ to $1.0$. As can be seen in Figure~\ref{fig:dataset}, when $\lambda$ gets larger, the AIWLS model fits the test set better.

We ran $1,000$ experiments to compute the risk estimated by different methods. The mean and standard deviation of the estimation are plotted in Figure~\ref{fig:means_and_stds}. Source Risk tends to deviate from Target Risk and is not a reasonable estimator. DEV and IWCV both correspond well with Target Risk, but after a closer look at Figure~\ref{fig:stds_in_detail}, we observe that DEV shows significantly smaller variance compared to IWCV, which justifies the efficacy of the control variate method. 

\subsection{\textbf{VisDA} Dataset}

\textbf{VisDA}~\cite{peng_visda} is a large-scale cross-domain dataset designed for domain adaptation in computer vision. The source domain in {VisDA} consists of synthetic rendered images, while the target domain images are cropped from either Microsoft COCO dataset~\cite{cite:ECCV14COCO} or Youtube Bounding Boxes dataset~\cite{cite:CVPR17BBox}.

We choose a state of the art model on the {VisDA} dataset, Maximum Classifier Discrepancy (\textbf{MCD})~\cite{cite:CVPR18MCD}, to explore the model selection efficacy of DEV. A key hyperparameter in MCD is \textit{the number of generator update iterations}, denoted as $k$. We try $k= 1, 2, 3, 4, 5$ together with various learning rates to select the best model.

Note that we tune MCD in terms of its mean accuracy, not by tuning each class and aggregating the best results. So the accuracy of Target Risk is only an upper bound with respect to the mean accuracy. According to Table~\ref{tab:MCD}, IWCV does not work well on {VisDA}. IWCV relies on pre-trained features to compute the density ratio estimation, but the domain gap in pre-trained features is large, leading to unstable importance weights and degraded performance of IWCV. In contrast, DEV successfully selects out the best model by working on the adapted features which endow smaller domain gap than the pre-trained features. According to the importance weighting theory in terms of R\'{e}nyi divergence \cite{cite:NIPS10LB}, this implies a bounded variance of DEV.

\subsection{\textbf{Office-31} Dataset}

\textbf{Office-31}~\cite{cite:ECCV10Office} is a standard dataset for visual domain adaptation. With $4652$ images in total, it is divided into $3$ domains: images downloaded from \textbf{A}mazon, photos taken by \textbf{D}SLR and \textbf{W}eb camera. Since photos from W and D contain the same objects, they are visually similar and present small domain gap. Images from A are usually dissimilar with images from W and D. 

We select another state of the art Deep UDA model on the {Office-31} dataset, Conditional Domain Adversarial Network (\textbf{CDAN})~\cite{cite:NIPS18CDAN} for evaluating model selection performance of different validation methods. In CDAN, an important hyperparameter is the trade-off coefficient $\lambda$, which balances between the transferability and the discriminability of the learned representations. We implement several trade-offs ($\lambda = 0.5, 0.75, 1.0, 1.25, 1.5$, with $\lambda = 1$ as its default setting) along with several learning rate configurations. Results are reported in Table~\ref{tab:cdan}. Performance tuned by Target Risk is the upper bound and we are glad to observe that DEV performs nearly as well as Target Risk, surpassing IWCV, Source Risk and results in the original papers. We observe that the selected model according to Source Risk only works well when the domain gap is pretty small as in the cases of D $\rightarrow$ W and W $\rightarrow$ D. By contrast, DEV works quite well even when the domain gap is large as in D $\rightarrow$ A.

\subsection{\textbf{Digits} Dataset}

\label{sec:Digits}

\textbf{Digits} \cite{cite:ICML15RevGrad} dataset consists of three domains: MNIST, USPS and SVHN. Among them, USPS and MNIST are similar, with white digits written by hand on black backgrounds. SVHN, however, is cropped from real street view images and introduces rich background noise. 

A state of the art generative model for domain adaptation on the {Digits} dataset is Generate to Adapt (\textbf{GTA})~\cite{cite:CVPR18GenerateAdapt}. We test the model selection performance of DEV on GTA in Table~\ref{tab:digit}. Besides the learning rate, we also tune the hyperparameters $\alpha$ and $\beta$ of GTA. Without altering the GTA model, DEV improves GTA (Original) by $1.4\%$. The model selected by Source Risk works poorly on task SVHN $\rightarrow$ MNIST. SVHN is visually very different from MNIST. The large domain gap makes the source risk deviate far from the target risk, resulting in inaccurate model selection. By contrast, even under large domain gap, DEV can exploit adapted features for accurate model selection.

\subsection{Beyond Standard Domain Adaptation}

\label{sec:go_beyond}
Standard domain adaptation relies on the assumption that source domain and target domain share the same label set. Several recent works relax the assumption and propose new variants of domain adaptation: Partial Domain Adaptation (PDA)~\cite{cite:CVPR18SAN, cite:ECCV18PADA} and Open Set Domain Adaptation (OSDA)~\cite{cite:ICCV17OpenSet,cite:ECCV18OpenSet}. 
In partial domain adaptation, the source label set subsumes the target label set, which naturally satisfies the assumption of DEV: $\textbf{supp} \; p_{{\bm f}} \supset \textbf{supp} \; q_{{\bm f}} $, where $p_{{\bm f}}$ and $q_{{\bm f}}$ are the adapted feature distributions. It is interesting to find out that DEV can handle model selection problems in partial domain adaptation without any modification.

We justify this by choosing a state of the art method, Partial Adversarial Domain Adaptation (\textbf{PADA})~\cite{cite:ECCV18PADA} and tuning it with DEV. Besides the learning rate, we also tune the hyperparameter \textit{update\_iteration} of PADA in $\left\{ 300, 400, 500, 600, 700\right\}$ (with $500$ as its default setting). The results are shown in Table~\ref{tab:PADA}. DEV continues to correspond well with Target Risk, even exceeding the original results reported in \cite{cite:ECCV18PADA}.

\begin{figure}[h]
	\includegraphics[width=0.95\columnwidth]{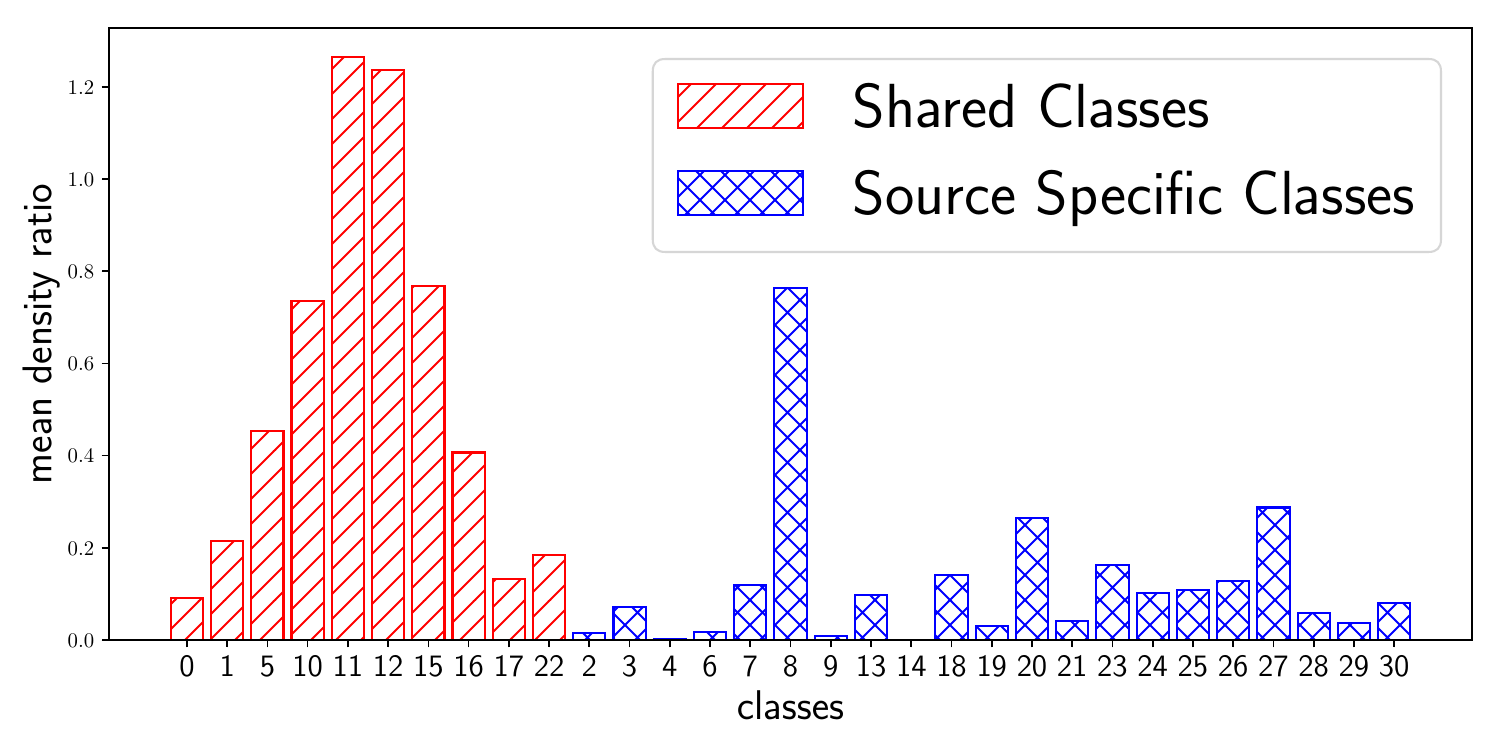}
		\vspace{-10pt}
	\caption{Estimated density ratio averaged across source classes on {partial domain adaptation} ~\cite{cite:ECCV18PADA} task A $\rightarrow$ W. Shared classes are marked in red while the other classes marked in blue. }
		\label{fig:weight_analysis}
\end{figure}

We further analyze the estimated density ratio in Figure~\ref{fig:weight_analysis}. Estimating the ground truth density ratio in practical datasets is intractable. However, in partial domain adaptation, $w_{{\bm f}}(\mathbf{x})$ for samples associated with source-specific classes should be smaller than for those of the shared classes. We plot the mean density ratio for each class,  marking the shared classes in red and the source-specific classes in blue, which justifies the assumption. It seems counterintuitive that the source-specific class $8$ has too high density ratio. By zooming in the dataset, we find that class $8$ is ``\textit{desktop\_computer}'', which is often confused with the shared classes ``\textit{laptop\_computer}'' and ``\textit{monitor}''. In summary, the density ratio estimated by DEV is generally accurate, which enables unbiased estimate of the target risk under controlled variance.

\subsection{Ablation Study}

To disentangle the contributions behind the success of DEV, we conduct an ablation study on VisDA dataset as shown in Table~\ref{tab:MCD}. The observation that DEV without control variate is superior to IWCV implies that model selection in Deep UDA can benefit largely from adapting feature representations. By plugging in the control variate method, DEV is further improved by over $2\%$, indicating the importance of variance control towards an accurate model selection in Deep UDA.

\section{Conclusion}

This paper introduced Deep Embedded Validation (DEV), an accurate model selection method in Deep UDA. DEV embeds deep adapted representations into the validation procedure to yield more reliable density ratio estimate, and leverages the control variate method to reduce the variance. Theoretical analysis and extensive experiments justify that DEV performs nearly on par with the Target Risk, significantly surpassing the previous methods. The superiority of DEV makes it an accurate, non-intrusive model selection method in the absence of labeled data in the target domain. 

\section*{Acknowledgements}
This work was supported by the National Natural Science Foundation of China (61772299, 71690231, and 61672313).

\bibliography{DEV}
\bibliographystyle{icml2019}

\end{document}